\documentclass{sig-alternate} %
\usepackage{graphicx}
\usepackage{subfigure}
\usepackage{times}
\usepackage{color}

\usepackage{algorithm}
\usepackage{algorithmic}

\usepackage{amssymb}
\usepackage{amssymb}
\usepackage{amsmath}
\usepackage{textcomp}

\usepackage{dblfloatfix}
\usepackage{xcolor}

\usepackage{bm}
\usepackage{dsfont}
\usepackage{url}
\usepackage{multirow}
\usepackage{microtype}
\usepackage{booktabs}
\usepackage{lmodern}
\usepackage[numbers]{natbib}

\newcommand{\argmin}{\operatornamewithlimits{argmin}}

\newcommand{\hide}[1]{}
\newcommand{\note}[1]{}

\newcommand{\vlambda}{{\bm{\lambda}}}
\newcommand{\vLambda}{{\bm{\Lambda}}}
\newcommand{\vs}[0]{\emph{vs}}
\newcommand{\eg}[0]{\emph{e.{}g.{}}}

\newcommand{\gauss}{\mbox{${\cal N}$}}

\newcommand{\interrowspace}{.6em}

\usepackage{etoolbox}

\begin{document}
\makeatletter
\def\@copyrightspace{\relax}
\makeatother

\title{ Auto-WEKA: Combined Selection and Hyperparameter Optimization of Classification Algorithms }

\author{Chris Thornton ~~~~~ Frank Hutter ~~~~~ Holger H. Hoos ~~~~~ Kevin Leyton-Brown\\~\\
\affaddr{Department of Computer Science, University of British Columbia}\\
\affaddr{201-2366 Main Mall, Vancouver BC, V6T 1Z4, Canada}\\
       \email{\{cwthornt, hutter, hoos, kevinlb\}@cs.ubc.ca}
}

\maketitle

\begin{abstract}
Many different machine learning algorithms exist; taking into account each algorithm's hyperparameters, there is a staggeringly large number of possible alternatives overall.  We consider the problem of simultaneously selecting a learning algorithm and setting its hyperparameters, going beyond previous work that addresses these issues in isolation.
We show that this problem can be addressed by a fully automated approach, leveraging recent innovations in Bayesian optimization.
Specifically, we consider a wide range of feature selection techniques (combining 3 search and 8 evaluator methods) and all classification approaches implemented in WEKA, spanning 2 ensemble methods, 10 meta-methods, 27 base classifiers, and hyperparameter settings for each classifier.
On each of 21 popular datasets from the UCI repository, the KDD Cup 09, variants of the MNIST dataset and CIFAR-10, we show classification performance
often much better than using standard selection/hyperparameter optimization methods.
We hope that our approach will help non-expert users to more effectively identify machine learning algorithms and hyperparameter settings appropriate to their applications, and hence to achieve improved performance.
\end{abstract}


\section{Introduction}
Increasingly, users of machine learning tools are non-experts who require off-the-shelf solutions.
The machine learning community has much aided such users by making available a wide variety of sophisticated learning algorithms and feature selection methods through open source packages, such as WEKA~\cite{hall2009weka} and PyBrain~\cite{schaul2010pybrain}. Each of these packages asks a user to make two kinds of choices: selecting a learning algorithm and customizing it by setting 
its hyperparameters (which also control any feature selection being performed).
It can be challenging to make the right choice when faced with 
these degrees of freedom, leaving many users to select algorithms based on reputation or intuitive appeal, and/or to leave hyperparameters set to default values.
Of course, this approach can yield performance far worse than that of the best method and hyperparameter settings.

This suggests a natural challenge for machine learning: given a dataset, to automatically and simultaneously choose a learning algorithm and set its hyperparameters to optimize empirical performance.
We dub this the \emph{combined algorithm selection and hyperparameter optimization problem} (short: CASH); we formally define it in Section \ref{sec:autoselect}. Despite the practical importance of this problem, we are surprised to find no evidence that it has previously been considered in the literature.
A likely explanation is that the combined space of learning algorithms and their hyperparameters is very challenging to search: the response function is noisy and the space is high dimensional, involves both categorical and continuous choices, and contains hierarchical dependencies (\eg, the hyperparameters of a learning algorithm are only meaningful if that algorithm is chosen; the algorithm choices in an ensemble method are only meaningful if that ensemble method is chosen; etc). In contrast, we do note that there has been considerable past work separately addressing model selection \citep[\eg{},][]{adankon2009model,biem2003model,bozdogan1987model,brazdil2003ranking,chapelle:2001,maron1994hoeffding,mcquarrie1998regression,Zhao:model_sel_lasso} and hyperparameter optimization \citep[\eg{},][]{bengio2000gradient,bergstra2011hyper,bergstra2012random,guo2008novelpso,snoek2012practical,strijov2010nonlinear}.

In what follows, we demonstrate that {CASH can be viewed as a single hierarchical hyperparameter optimization problem, in which even the choice of algorithm itself is considered a hyperparameter. We also show that --- based on this problem formulation --- recent Bayesian optimization methods can obtain high quality results in reasonable time and with minimal human effort.}
After discussing some preliminaries (Section \ref{sec:prelim}), we define the CASH problem and discuss methods for tackling it (Section~\ref{sec:autoselect}).
We then define a concrete {CASH}
problem encompassing the full range of classifiers and feature selectors in the open source package WEKA (Section~\ref{sec:autoweka}), and show that a search in the combined space of algorithms and hyperparameters yields better-performing models {than standard algorithm selection/hyperparameter optimization methods (Section~\ref{sec:eval})}. 
More specifically, we show that the {recent} Bayesian optimization procedures TPE~\cite{bergstra2011hyper} and SMAC~\cite{hutter2011smac} find combinations of algorithms and hyperparameters that
often outperform existing baseline methods, especially on large datasets.

\section{Preliminaries}\label{sec:prelim}
This work focuses on classification problems: learning a function $f: \mathcal{X} \mapsto \mathcal{Y}$ with finite $\mathcal{Y}$.
A \emph{learning algorithm} $A$ maps a set $\{d_1,\dots,d_n\}$ of training data points $d_i = (\mathbf{x}_i,y_i) \in \mathcal{X} \times \mathcal{Y}$ to such a function, which is often expressed via a vector of \emph{model parameters}.
Most learning algorithms $A$ further expose \emph{hyperparameters} $\vlambda \in \vLambda$, which change the way the learning algorithm $A_\vlambda{}$ itself works. For example, hyperparameters are used to describe a description-length penalty, the number of neurons in a hidden layer, the number of data points that a leaf in a decision tree must contain to be eligible for splitting, etc.
These hyperparameters are typically optimized in an ``outer loop'' that evaluates the performance of each hyperparameter configuration using cross-validation.

\subsection{Model Selection}\label{sec:model_sel}

Given a set of learning algorithms $\mathcal{A}$ and a limited amount of training data $\mathcal{D} = \{(\mathbf{x}_1,y_1),\dots,(\mathbf{x}_n,y_n)\}$,
the goal of model selection is to determine the algorithm $A^* \in \mathcal{A}$ with optimal generalization performance.
Generalization performance is estimated by splitting $\mathcal{D}$ into disjoint training and validation sets $\mathcal{D}_{\text{train}}^{(i)}$ and $\mathcal{D}_{\text{valid}}^{(i)}$, learning functions $f_i$ by applying $A^*$ to $\mathcal{D}_{\text{train}}^{(i)}$, and evaluating the predictive performance of these functions on $\mathcal{D}_{\text{valid}}^{(i)}$.
This allows for the model selection problem to be written as:
\vspace*{-0.2cm}
\begin{eqnarray}
\label{hyperparamopt}
\nonumber{}A^* \in \argmin_{A \in \mathcal{A}} \frac{1}{k} \sum_{i=1}^{k} \mathcal{L}(A, \mathcal{D}_{\text{train}}^{(i)}, \mathcal{D}_{\text{valid}}^{(i)}),
\end{eqnarray}
\vspace*{-0.2cm}

\noindent{}where $\mathcal{L}(A, \mathcal{D}_{\text{train}}^{(i)}, \mathcal{D}_{\text{valid}}^{(i)})$ is the
loss (here: misclassification rate) achieved by $A$ when trained on $\mathcal{D}_{\text{train}}^{(i)}$ and evaluated on $\mathcal{D}_{\text{valid}}^{(i)}$.
{We use $k$-fold cross-validation~\cite{kohavi1995study}, which splits the training data into $k$ equal-sized partitions
$\mathcal{D}_{\text{valid}}^{(1)}, \dots, \mathcal{D}_{\text{valid}}^{(k)}$, and sets $\mathcal{D}_{\text{train}}^{(i)} = \mathcal{D} \setminus{} \mathcal{D}_{\text{valid}}^{(i)}$ for $i=1,\dots,k$.}\footnote{K-fold cross-validation is not the only available method for estimating generalization performance. We also experimented with the technique of repeated random subsampling validation~\cite{kohavi1995study}, with similar results.}

\subsection{Hyperparameter Optimization}\label{sec:hyp_opt}

The problem of optimizing the hyperparameters $\vlambda \in \vLambda$ of a given learning algorithm $A$ is conceptually similar to that of model selection. Some key differences are that hyperparameters are often continuous, that hyperparameter spaces are often high dimensional, and that we can exploit correlation structure between different hyperparameter settings $\vlambda_1, \vlambda_2 \in \vLambda$.
Given $n$ hyperparameters $\lambda_1, \dots, \lambda_n$ with domains $\Lambda_1, \dots, \Lambda_n$, the hyperparameter space $\vLambda$ is a subset of the crossproduct
of these domains: $\vLambda \subset \Lambda_1 \times \dots \times \Lambda_n$.
This subset is often strict, such as when certain settings of one hyperparameter render other hyperparameters inactive. For example,
the parameters determining the specifics of the third layer of a deep belief network are not relevant if
the network depth is set to one or two. Likewise, the parameters of a support vector machine's polynomial kernel
are not relevant if we use a different kernel instead.

More formally, following~\cite{hutter2009paramils}, we say that a hyperparameter $\lambda_i$ is \emph{conditional}
on another hyperparameter $\lambda_j$, if $\lambda_i$ is only active if
hyperparameter $\lambda_j$ takes values from a given set $V_i(j) \subsetneq \Lambda_j$;
in this case 
we call $\lambda_j$ a \emph{parent} of $\lambda_i$.
Conditional hyperparameters can in turn be parents of other conditional hyperparameters, giving rise to a
tree-structured space~\cite{bergstra2011hyper} or, in some cases, a directed acyclic graph (DAG)~\cite{hutter2009paramils}.
Given such a structured space $\vLambda$,
the {(hierarchical)} hyperparameter optimization problem can be written as:
\vspace*{-0.2cm}
\begin{eqnarray}
\nonumber{}\vlambda^* \in \argmin_{\vlambda \in \vLambda} \frac{1}{k} \sum_{i=1}^{k} \mathcal{L}(A_{\vlambda}, \mathcal{D}_{\text{train}}^{(i)}, \mathcal{D}_{\text{valid}}^{(i)}).
\end{eqnarray}

\section{Combined Algorithm Selection and Hyperparameter Optimization (CASH)}\label{sec:autoselect}

Given a set of algorithms $\mathcal{A} = \{A^{(1)}, \dots, A^{(k)}\}$ with associated hyperparameter spaces $\vLambda^{(1)}, \dots, \vLambda^{(k)}$, we define
the combined algorithm selection and hyperparameter optimization problem (CASH) as computing
\vspace*{-0.2cm}
\begin{eqnarray}
\label{eq:CAHS}
{A^*}_{\vlambda^*} \in \argmin_{A^{(j)} \in \mathcal{A}, \vlambda \in \vLambda^{(j)}} \frac{1}{k}  \sum_{i=1}^{k} \mathcal{L}(A^{(j)}_\vlambda, \mathcal{D}_{\text{train}}^{(i)}, \mathcal{D}_{\text{valid}}^{(i)}).
\end{eqnarray}
\vspace*{-0.2cm}

{We note that this problem can be reformulated as a single combined hierarchical hyperparameter optimization problem}
with parameter space $\vLambda = \vLambda^{(1)} \cup \cdots \cup \vLambda^{(k)} \cup \{ \lambda_r \}$, where
$\lambda_r$ is a new root-level hyperparameter that selects between algorithms $A^{(1)}, \dots, A^{(k)}$.
The root-level parameters of each subspace $\vLambda^{(i)}$ are made conditional on $\lambda_r$ being instantiated to $A_i$.

In principle, Problem \ref{eq:CAHS} can be tackled in various ways.
A promising approach is Bayesian Optimization~\cite{Brochu:2009}, and in particular Sequential Model-Based Optimization \citep[SMBO;][]{hutter2011smac}, a versatile stochastic optimization framework that can work explicitly with both categorical and continuous hyperparameters, and that can exploit hierarchical structure stemming from conditional parameters.
{SMBO (outlined in Algorithm~\ref{alg:smbo}) first builds a model $\mathcal{M}_\mathcal{L}$ that captures the dependence of
loss function $\mathcal{L}$ on hyperparameter settings $\vlambda$ (line 1 in Algorithm~\ref{alg:smbo}). It then iterates the following steps:
use $\mathcal{M}_\mathcal{L}$ to determine a promising candidate configuration of hyperparameters $\vlambda$ to evaluate next (line 3); 
evaluate the loss $c$ of $\vlambda$ (line 4); and update the model $\mathcal{M}_\mathcal{L}$ with the new data point $(\vlambda,c)$ thus obtained (lines 5--6).}

\begin{algorithm}[t]
{\footnotesize
\caption{SMBO}
\label{alg:smbo}
\begin{algorithmic}[1]
    \label{line:init}\STATE initialise model $\mathcal{M}_L$;  $\mathcal{H} \gets \emptyset$
    \WHILE{time budget for optimization has not been exhausted}
      \label{line:get_lambda}\STATE $\vlambda \gets$ candidate configuration from $\mathcal{M}_L$
      \label{line:get_c}\STATE Compute $c = \mathcal{L}(A_{\vlambda}, \mathcal{D}_{\text{train}}^{(i)}, \mathcal{D}_{\text{valid}}^{(i)})$
      \label{line:update_H}\STATE $\mathcal{H} \gets \mathcal{H} \cup \left\{(\vlambda,c)\right\}$
      \label{line:update_M}\STATE Update $\mathcal{M}_L$ given $\mathcal{H}$
    \ENDWHILE
    \STATE \textbf{return} $\vlambda$ from $\mathcal{H}$ with minimal $c$
\end{algorithmic}
}
\end{algorithm}

{In order to select its next hyperparameter configuration $\vlambda$ using model $\mathcal{M}_{\mathcal{L}}$, SMBO uses a so-called \emph{acquisition function} $a_{\mathcal{M}_{\mathcal{L}}}:\vLambda \rightarrow \mathds{R}$, which uses the predictive distribution of model $\mathcal{M}_{\mathcal{L}}$ at arbitrary hyperparameter configurations $\vlambda \in \vLambda$ to quantify (in closed form) how useful knowledge about $\vlambda$ would be. SMBO then simply maximizes this function over $\vLambda$ to select the most useful configuration $\vlambda$ to evaluate next. Several prominent acquisition functions exist~\citep{JonSchWel98,SchWelJon98,SriEtAl10:GP-UCB} that all aim to automatically trade off exploitation (locally optimizing hyperparameters in regions known to perform well) versus exploration (trying hyperparameters in a relatively unexplored region of the space) in order to avoid premature convergence. In this work, we use one of the most prominent acquisition functions, the \emph{positive expected improvement (EI)} attainable over an existing given error rate $c_{min}$~\citep{SchWelJon98}. Let $c(\vlambda)$ denote the error rate of hyperparameter configuration $\vlambda$. Then, the positive improvement function over $c_{min}$ is defined as:
\[I_{c_{min}}(\vlambda) := \max\{c_{min}-c(\vlambda),0\}.\]
Of course, we do not know $c(\vlambda)$. We can, however, compute its expectation with respect to the current model $\mathcal{M}_{\mathcal{L}}$:
\begin{equation}
\label{eqn:ei}\mathds{E}_{\mathcal{M}_{\mathcal{L}}}[I_{c_{min}}(\vlambda{})] := \int_{-\infty}^{c_{min}} \max\{c_{min}-c,0\}\cdot p_{\mathcal{M}_{L}}(c \mid \lambda) \; dc.
\end{equation}
One main difference between existing SMBO algorithms lies in the model class they employ.
We now review the two SMBO algorithms whose models can handle hierarchical hyperparameters and that are thus suitable for the task of combined algorithm 
selection and hyperparameter optimization.}

\subsection{Sequential Model-based Algorithm Configuration (SMAC)} \label{sec:smac}

Sequential model-based algorithm configuration \citep[SMAC;][]{hutter2011smac} 
supports a variety of models $p(c \mid \vlambda)$ to capture the dependence of the loss function $c$ on hyper-parameters $\vlambda$,
including approximate Gaussian processes and random forests.
In this paper we use random forest models, since they tend to perform well with discrete and high-dimensional input data. 
SMAC handles conditional parameters by instantiating inactive conditional parameters in $\vlambda$ to default values for model training and prediction. {This allows the individual decision trees to include splits of the kind ``is hyperparameter $\lambda_i$ active?'', allowing them to focus on active hyperparameters.}
While random forests are not usually treated as probabilistic models, SMAC obtains a predictive mean $\mu_{\vlambda}$ and variance ${\sigma_{\vlambda}}^2$ of $p(c \mid \vlambda)$ as frequentist estimates over the predictions of its individual trees for $\vlambda$; it then models $p_{\mathcal{M}_{\mathcal{L}}}(c \mid \vlambda)$ as a Gaussian $\gauss(\mu_\vlambda, {\sigma_\vlambda}^2)$.

{SMAC uses the expected improvement criterion defined in Equation \ref{eqn:ei}, instantiating $c_{min}$ to the error rate of the best hyperparameter configuration measured so far. Under SMAC's predictive distribution
$p_{\mathcal{M}_{\mathcal{L}}}(c \mid \vlambda) = \gauss(\mu_\vlambda, {\sigma_\vlambda}^2)$, this expectation can be computed by the closed-form expression
\vspace*{-0.1cm}
\begin{equation}
\nonumber
\mathds{E}_{\mathcal{M}_{\mathcal{L}}}[I_{c_{min}}(\vlambda{})] = 
\sigma_{\vlambda} \cdot [u \cdot \Phi(u) + \varphi(u)]
\label{exp-imp-closed},
\end{equation}
where $u = \frac{c_{\min}-\mu_{\vlambda}}{\sigma_{\vlambda}}$, and $\varphi$ and $\Phi$ denote the probability density
function and cumulative distribution function of a standard normal distribution, respectively~\cite{JonSchWel98}.}

SMAC is designed for robust optimization under noisy function evaluations, and as such implements special mechanisms to keep track of its best known configuration and assure high confidence in its estimate of that configuration's performance. This robustness against noisy function evaluations can be exploited in combined algorithm selection and hyperparameter optimization, since the function to be optimized in Equation~\eqref{hyperparamopt} is a mean over a set of loss terms (each corresponding to one pair of $\mathcal{D}_{\text{train}}^{(i)}$ and $\mathcal{D}_{\text{valid}}^{(i)}$ constructed from the training set). A key idea in SMAC is to make progressively better estimates of this mean by evaluating these terms one at a time, thus trading off accuracy against computational cost.
In order for a new configuration to become a new incumbent, it must outperform the previous incumbent in every comparison made: considering only one fold, two folds, and so on up to the total number of folds previously used to evaluate the incumbent. (Furthermore, every time the incumbent survives such a comparison, it is evaluated on a new fold, up to the total number available, meaning that the number of folds used to evaluate the incumbent grows over time.) This means that a poorly performing configuration can be discarded after considering as little as a single fold.

Finally, SMAC also implements a diversification mechanism to achieve robust performance even when its model is misled: every second configuration is selected at random. Because of the evaluation procedure just described, the overhead required by this safeguard is less than it might appear.

\subsection{Tree-structured Parzen Estimator (TPE)}\label{sec:tpe}

{While SMAC models $p(c \mid \vlambda{})$ explicitly, the Tree-structure Parzen Estimator \citep[TPE;][]{bergstra2011hyper} uses separate models for $p(c)$ and $p(\vlambda{} \mid c)$. Specifically, it models $p(\vlambda{} \mid c)$ as one of two density estimates, conditional on whether $c$ is greater or less than a given threshold value $c^*$:
\vspace*{-0.3cm}
\begin{equation}
  p(\vlambda{} \mid c)=\begin{cases}
    \ell(\vlambda), & \text{if $c<c^*$}.\\
    g(\vlambda), & \text{if $c\ge c^*$}.
  \end{cases}
\vspace*{-0.2cm}
\end{equation}
Here, $c^*$ is chosen as the $\gamma$-quantile of the losses TPE obtained so far (where $\gamma$ is an algorithm parameter with a default value of $\gamma=0.15$), $\ell(\cdot)$ is a density estimate learned from all previous hyperparameters $\vlambda$ with corresponding loss smaller than $c^*$, and $g(\cdot)$ is a density estimate learned from all previous hyperparameters $\vlambda$ with corresponding loss greater than or equal to $c^*$.}
Intuitively, this creates a probabilistic density estimator $\ell(\cdot)$ for hyperparameters that appear to do `well', and a different density estimator $g(\cdot)$ for hyperparameters that appear `poor' with respect to the threshold. {Bergstra et al.~\cite{bergstra2011hyper} showed that the expected improvement $\mathds{E}_{\mathcal{M}_{\mathcal{L}}}[I_{c_{min}}(\vlambda{})]$ from Equation \ref{eqn:ei} is proportional to a quantity that can be computed in closed-form from $\gamma$, $g(\vlambda{})$, and $\ell(\vlambda{})$:
\[\mathds{E}[I_{c_{min}}(\vlambda{})] \propto \left(\gamma + \frac{g(\vlambda{})}{\ell(\vlambda{})}  \cdot (1-\gamma)\right)^{-1}.\] 
TPE maximizes this expression by generating many candidate hyperparameter configurations at random and picking $\vlambda{}$ with the smallest value of $g(\vlambda{})/\ell(\vlambda{})$.}

{The density estimators $\ell(\cdot)$ and $g(\cdot)$ have a hierarchical structure
with discrete, continuous, and conditional variables reflecting the hyperparameters and their dependence relationships.
For each node in this tree structure}, a 1-D Parzen estimator is created to model the density of the node's corresponding hyperparameter. {For a given hyperparameter configuration $\vlambda{}$ that is being added to either $\ell$ or $g$, only the 1-D estimators corresponding to active hyperparameters in $\vlambda{}$ are updated.
For continuous hyperparameters, these 1-D estimators are constructed by placing density in the form of a Gaussian at each hyperparameter value $\vlambda{}_i$, with standard deviation set to the larger of each point's left and right neighbour. Discrete hyperparameters are estimated with probabilities proportional to the number of times that a particular choice occurred in the set of observations.
To evaluate a candidate hyperparameter $\vlambda{}$'s probability estimate, TPE starts at the root of the tree and descends into the leaves by following paths that only use active hyperparameters. At each node in this traversal, the probability of the corresponding hyperparameter is computed according to its 1-D estimator, and the individual probabilities are combined on a pass back up to the root of the tree.}
{Note that this means that TPE assumes independence for hyperparameters that do not appear together along any path from the tree's root to one of its leaves.}
\section{Auto-WEKA}\label{sec:autoweka}
\begin{table}[t]
\vskip -0.05in
  \caption{Classifiers in Auto-WEKA: Classifiers marked with~$^*$ are meta-methods, which in addition to their own parameters take one `base' classifier and its parameters. Classifiers marked with~$^+$ are ensemble methods that take as input up to 5 `base' classifiers and their parameters. \emph{Categorical} and \emph{Numeric} refer to the number of hyperparameters of each kind for each classifier.}
  \label{table:classifiers}
  \begin{center}
  \begin{sc}
    \setlength{\tabcolsep}{1.8pt}
    {\scriptsize\centering
      \begin{tabular}{l@{\hskip .6em}cc}\toprule
      \textbf{Classifier} &\textbf{Categorical} &\textbf{Numeric} \\
      \midrule
Bayes Net & 2 & 0 \\
Naive Bayes & 2 & 0 \\
Naive Bayes Multinomial & 0 & 0 \\
Gaussian Process & 3 & 6 \\
Linear Regression & 2 & 1 \\
Logistic Regression& 0 & 1 \\
Single-Layer Perceptron & 5 & 2 \\
Stochastic Gradient Descent & 3 & 2 \\
SVM & 4 & 6 \\
Simple Linear Regression & 0 & 0 \\
Simple Logistic Regression & 2 & 1 \\
Voted Perceptron & 1 & 2 \\
KNN & 4 & 1 \\
K-Star & 2 & 1 \\
Decision Table & 4 & 0 \\
RIPPER & 3 & 1 \\
M5 Rules & 3 & 1 \\
1-R & 0 & 1 \\
PART & 2 & 2 \\
0-R & 0 & 0 \\
Decision Stump & 0 & 0 \\
C4.5 Decision Tree & 6 & 2 \\
Logistic Model Tree & 5 & 2 \\
M5 Tree & 3 & 1 \\
Random Forest & 2 & 3 \\
Random Tree & 4 & 4 \\
REP Tree & 2 & 3 \\
\addlinespace[\interrowspace]
Locally Weighted Learning$^*$ & 3 & 0 \\
AdaBoost M1$^*$ & 2 & 2 \\
Additive Regression$^*$ & 1 & 2 \\
Attribute Selected$^*$ & 2 & 0 \\
Bagging$^*$ & 1 & 2 \\
Classification via Regression$^*$ & 0 & 0 \\
LogitBoost$^*$ & 4 & 4 \\
MultiClass Classifier$^*$ & 3 & 0 \\
Random Committee$^*$ & 0 & 1 \\
Random Subspace$^*$ & 0 & 2 \\
\addlinespace[\interrowspace]
Voting$^+$ & 1 & 0 \\
Stacking$^+$ & 0 & 0 \\
      \bottomrule
    \end{tabular}
    }
    \end{sc}
  \end{center}
  \vskip -0.23in
\end{table}

\begin{table}[t]
\vskip -0.05in  
\caption{Feature Search/Evaluator methods in Auto-WEKA: Methods marked with~$^*$ are search methods, which require one feature evaluator that is used to determine the importance of a feature. \emph{Categorical} and \emph{Numeric} refer to the number of hyperparameters of each method.}
  \label{table:featselection}
  \begin{center}
  \begin{sc}
    \setlength{\tabcolsep}{1.8pt}
    {\scriptsize\centering
      \begin{tabular}{l@{\hskip .6em}cc}\toprule
      \textbf{Feature Method} &\textbf{Categorical} &\textbf{Numeric} \\
      \midrule
\addlinespace[\interrowspace]
Best First$^*$ & 1 & 1 \\
Greedy Stepwise$^*$ & 3 & 2 \\
Ranker$^*$ & 0 & 1 \\
\addlinespace[\interrowspace]
CFS Subset Eval & 2 & 0 \\
Pearson Correlation Eval & 0 & 0 \\
Gain Ratio Eval & 0 & 0 \\
Info Gain Eval & 2 & 0 \\
1-R Eval & 1 & 2 \\
Principal Components Eval & 2 & 2 \\
RELIEF Eval & 1 & 2 \\
Symmetrical Uncertainty  Eval & 1 & 0 \\
      \bottomrule
    \end{tabular}
    }
    \end{sc}
  \end{center}
  \vskip -0.23in
\end{table}

To demonstrate the feasibility of an automatic approach to solving the CASH problem, we built a tool, \emph{Auto-WEKA}, that solves this
problem for all classification algorithms and feature selectors/evaluators implemented in the WEKA package \cite{hall2009weka}.
Note that while we have focused on classification algorithms in WEKA, there is no obstacle to extending our approach to other settings.

Table~\ref{table:classifiers} provides a list of all 39 WEKA classification
algorithms. Of these models, 27 are considered `base' classifiers (which
can be used independently), 10 of the remaining classifiers are meta
methods (which take a single base classifier and its parameters as an
input), and the final 2 ensemble classifiers can take any number of
base classifiers as input. We allowed the meta-methods to use any base
classifier with any hyperparameter settings, and allowed ensemble methods
to use up to five base classifiers, again with any hyperparameter settings.
{Not all classifiers are applicable on all datasets (e.g., due to a classifier's inability to handle missing data). For a given dataset, our Auto-WEKA implementation automatically only considers the subset of applicable classifiers.}

{Table~\ref{table:featselection} provides a list of WEKA's 3 feature search methods, as well
its 8 feature evaluators, and their respective number of subparameters (up to 5 for search; up to 4 for evaluators). 
To perform feature selection, a search method is combined with a feature evaluator, 
and the subparameters of both of them need to be instantiated. 
Feature selection is run as a pre-processing phase before building any classifier.}

The algorithms in Table~\ref{table:classifiers} and \ref{table:featselection} have a wide variety of
hyperparameters, which take on values from continuous intervals, from ranges of
integers, and from other discrete sets. We associated either a uniform or log
uniform prior with each numerical parameter, depending on its semantics. For
example, we set a log uniform prior for the ridge regression penalty, and a
uniform prior for the maximum depth for a tree in a random forest. Auto-WEKA works with continuous hyperparameter values directly; nevertheless, {to give a sense of the size of the hypothesis space we studied, we note that discretizing hyperparameter domains to a maximum of 10 values each would give rise to over $10^{47}$ hyperparameter settings. 
We emphasize that this space is \emph{much} larger than a simple union of the base learners' hypothesis spaces (whose size is roughly $10^8$), since the ensemble methods allow up to 5 \emph{independent} base learners, giving rise to a space with roughly $(10^8)^5 = 10^{40}$ elements. The feature selection part gives rise to another independent decision between roughly $10^{6}$ choices, and several parameters on the meta and ensemble level contribute another order of magnitude to the total size of AutoWEKA's hypothesis space.}

\begin{figure}[t]
\begin{center}
 \includegraphics[width=0.99\linewidth]{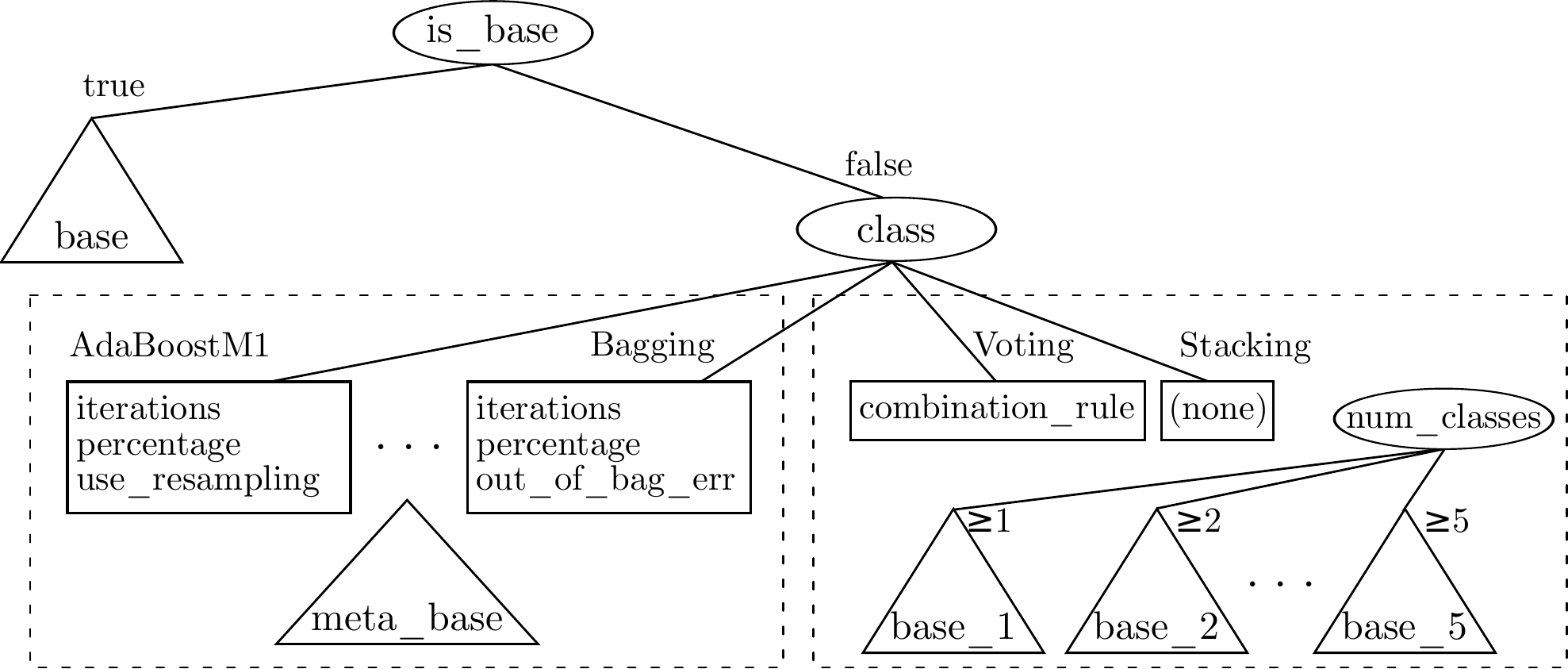}~\\~\\ 
 \includegraphics[width=0.99\linewidth]{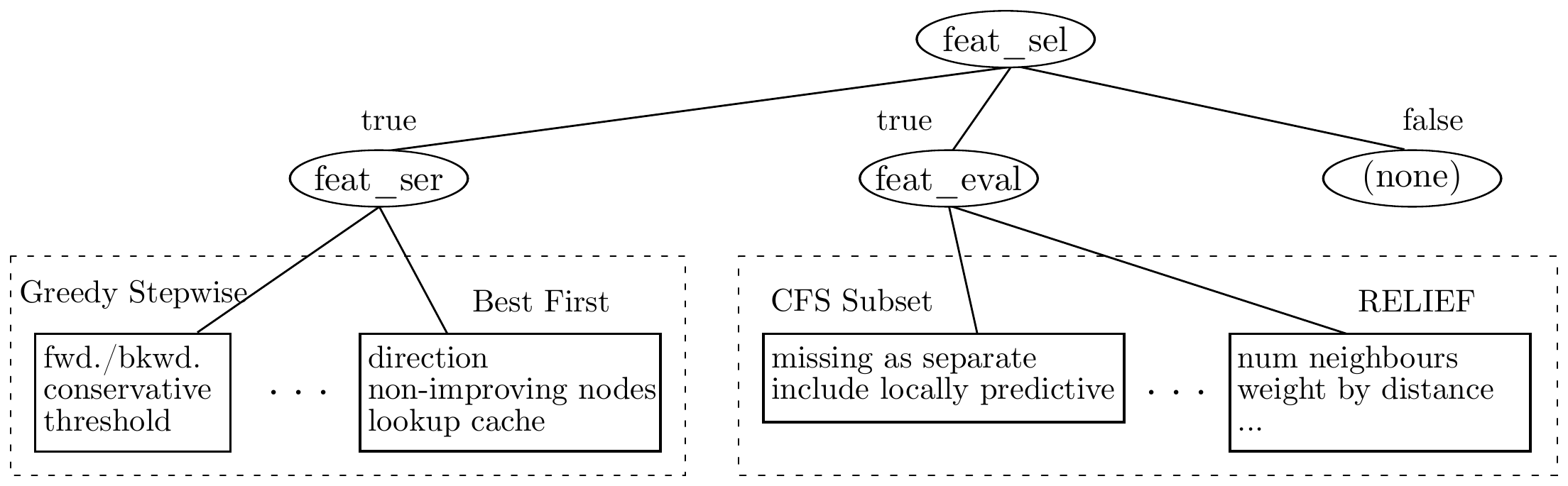}
 \end{center}
  \vspace*{-0.2in}
  \caption{Auto-WEKA's parameter space. Top: first top level Boolean, concerning Auto-WEKA's classification methods. The triangular items represent a parameter that selects one of the 27 base classifiers, and adds conditional classifier hyperparameters accordingly. Bottom: second top level Boolean, concerning Auto-WEKA's feature selection methods.}
 \label{fig:paramspace}
  \vspace*{-0.1in} 
\end{figure}

Auto-WEKA can be understood as a single learning algorithm with a highly conditional parameter space, as depicted in Figure~\ref{fig:paramspace}. {Auto-WEKA has two top-level Boolean parameters, the first of which is $\mathtt{is\_base}$} that selects among single
base classifiers and ensemble or meta-classifiers. If $\mathtt{is\_base}$
is $\mathtt{true}$, then the parameter $\mathtt{base}$ determines which of
the 27 base classifiers are to be used. If $\mathtt{is\_base}$ is
$\mathtt{false}$, then $\mathtt{class}$ indicates either an ensemble or a
meta-classifier. If $\mathtt{class}$ is a meta-classifier, then the
parameter $\mathtt{meta\_base}$ is chosen to be one of the 27 base
classifiers. In the event that $\mathtt{class}$ is an ensemble classifier,
an additional parameter $\mathtt{num\_classes}$ is an integer chosen from $\{1, \ldots, 5\}$. $\mathtt{base\_i}$ variables are then selected according to the value of
$\mathtt{num\_classes}$, which again select which of the 27 base
classifiers to use. For each $*\mathtt{base}*$ parameter, conditional
hyperparameters for every model are attached. 

{The second top level Boolean parameter $\mathtt{feat\_sel}$ indicates if one of the feature selection methods is going to be applied. If $\mathtt{feat\_sel}$ is $\mathtt{false}$, then Auto-WEKA passes the unmodified dataset to the classifier. If it is true, then $\mathtt{feat\_search}$ selects the choice of feature search method, and $\mathtt{feat\_eval}$ selects the choice of feature evaluator.}
This results in a very wide
tree that captures all the hierarchical nature of the model
hyperparameters, and allows the creation of a single hyperparameter
optimization problem with four hierarchical layers of a total of 786 parameters.

{Auto-WEKA is agnostic to the choice of optimizer, so we implemented variants leveraging  
SMAC and TPE, respectively -- the two Bayesian optimization algorithms that can handle hierarchical parameter spaces (see Sections \ref{sec:smac} and \ref{sec:tpe}.}\footnote{We thank the authors of SMAC and TPE for giving us access to their respective implementations.} 
We defined two Auto-WEKA variants, based on SMAC and TPE, respectively.
We made both of these Auto-WEKA versions available to the public at \url{www.cs.ubc.ca/labs/beta/Projects/autoweka/} and are committed to provide support for their widespread use in practice. 
{Both TPE and SMAC have their own parameters that influence their performance (such as TPE's choice of the $\gamma$-quantile indicating `good' or `bad' performance, or the parameters of SMAC's random forest model). In Auto-Weka, we used the defaults for these meta-hyperparameters, as set by the authors (we assume further small improvements may be possible by meta-hyperparameter optimization, but a separate process with a meta-training/validation set split would be required to guard against over-fitting, and we did not attempt this).}

Finally, we note that both TPE and SMAC are randomized algorithms and thus expected to produce different results based on the random seed provided. As demonstrated in~\cite{HutHooLey12-ParallelAC}, this allows for trivial, yet effective parallelization of the optimization process: simply perform $k$ independent runs of the optimization method in parallel and select the result of the run with the lowest cross-validation error to return.\footnote{{Other, more sophisticated methods for the parallelization of Bayesian optimization exist~\cite{HutHooLey12-ParallelAC,bergstra2011hyper,desautels12parallelizing,snoek2012practical}, but to date, there is no empirical evidence that these methods outperform the simple approach we use here when the cost of evaluating hyperparameter configurations varies across the space.}}
{We verified experimentally that the more parallel runs, the faster this process identifies high-quality configurations; nevertheless, we restricted the version of Auto-WEKA studied here to run only 4 parallel processes, in order to study a setting typical for commonly used workstations.} 

\section{Evaluating Auto-WEKA}\label{sec:eval}

We now describe an experimental study of the performance that can be achieved by
Auto-WEKA on various datasets. After specifying our experiment environment, we demonstrate the importance of addressing the algorithm selection and the CASH problems, and establish baselines for them (Section \ref{sec:baselines}). 
We then demonstrate Auto-WEKA's ability to search its enormous hyperparameter space effectively to find algorithms and hyperparameters
with low cross-validation error
(Section \ref{sec:exp_result_cv}). Then, we analyse its test performance and address concerns regarding overfitting (Section \ref{sec:exp_result_test}). Finally, we provide a synopsis of the classifiers and feature search/evaluators Auto-WEKA chose
in our experiments (Section \ref{sec:class_selected}).

\subsection{Experimental setup}

\begin{table}[t]
  \vskip -0.05in
  \caption{Datasets Used; {\em{Num.\ Discr.}}.\ and {\em{Num.\ Cont}}.\ refer to the number of discrete and continuous attributes of elements in the dataset, respectively.}
  \label{table:datasets}
\vspace*{-0.05in}
  \footnotesize{
  \begin{center}
\begin{sc}
  \setlength{\tabcolsep}{1.8pt}
    {\scriptsize
\centering
\begin{tabular}{l@{\hskip .6em}ccccc}
    \toprule

\multirow{2}{*}{\textbf{Name}} & \textbf{Num} & \textbf{Num} & \textbf{Num} & \textbf{Num} & \textbf{Num} \\
& \textbf{Discr.} & \textbf{Cont.} &\textbf{Classes} & \textbf{Training} & \textbf{Test} \\
\midrule

Dexter & $20\,000$ & 0 & 2& 420 & 180 \\
GermanCredit & 13 & 7 & 2 & 700 & 300 \\
Dorothea & $100\,000$ & 0 & 2 & 805 & 345\\
Yeast & 0 & 8 & 10 & $1\,038$ & 446  \\
Amazon & $10\,000$ & 0 & 49 & $1\,050$ & 450  \\
Secom & 0 & 591 & 2 & $1\,096$ & 471  \\
Semeion & 256 & 0 & 10 & $1\,115$ & 478  \\
Car & 6 & 0 & 4 & $1\,209$ & 519 \\
Madelon & 500 & 0 & 2 &$1\,820$ & 780 \\
KR-vs-KP & 37 & 0 & 2 & $2\,237$ & 959 \\
Abalone & 1 & 7 & 28 & $2\,923$ & $1\,254$  \\
Wine Quality & 0 & 11 & 11 & $3\,425$ & $1\,469$  \\
Waveform & 0 & 40 & 3 & $3\,500$ & $1\,500$  \\
Gisette & $5\,000$ & 0 & 2 & $4\,900$ & $2\,100$  \\
Convex & 0 & 784 & 2 & $8\,000$ & $50\,000$  \\
\addlinespace[\interrowspace]
CIFAR-10-Small & $3\,072$ & 0 & 10 & $10\,000$ &  $10\,000$ \\
MNIST Basic & 0 & 784 & 10 & $12\,000$ & $50\,000$  \\
Rot. MNIST + BI & 0 & 784 & 10 & $12\,000$ & $50\,000$  \\
Shuttle & 9 & 0 & 7 & $43\,500$ & $14\,500$  \\
KDD09-Appentency & 190 & 40 & 2 & $35\,000$ & $15\,000$ \\
CIFAR-10 & $3\,072$ & 0 & 10 & $50\,000$ &  $10\,000$ \\
\bottomrule

 \end{tabular}
 }
\end{sc}
  \end{center}
  }
  \vskip -0.23in
\end{table}

{We evaluated Auto-WEKA on 21 prominent benchmark datasets (see Table \ref{table:datasets}): 15 sets from the UCI repository~\cite{UCI-rep}; the `convex', `MNIST basic' and `rotated MNIST with background images' tasks used in \cite{bergstra2012random}; the appentency task from the KDD Cup '09; and two versions of the CIFAR-10 image classification task \cite{krizhevsky2009learning} (CIFAR-10-Small is a subset of CIFAR-10, where only the first $10\,000$ training data points are used rather than the full $50\,000$.)}
For datasets with a predefined training/test split, we used that split. Otherwise, we randomly split the dataset into 70\% training and 30\% test data. 
The test data was never seen by any optimization method; it was only used once in an offline analysis stage to evaluate the models found by the various optimization methods. {We denote datasets with at least $10\,000$ training data points as `large' and all others as `small'.}

All of our experiments were run on Linux machines with Intel Xeon X5650 six-core processors, running at 2.66GHz.
All datasets had a RAM limit of 3GB for classification; if training a classifier ever exceeded this memory limit, the classifier job was terminated, returning a misclassification rate of 100\%. An additional 1GB of RAM was allocated for the SMBO method. {While these limits are somewhat arbitrary, 
we believe them to be reasonably close to the resource limitations faced by any user of machine learning algorithms.}
We also limited the training time for each evaluation of a learning algorithm on each fold, to ensure that the optimization method had a chance to explore the search space. Once this training budget for a fold is consumed, Auto-WEKA sends an interrupt to the learning algorithm to terminate as soon as possible, and the (partially) trained model is then evaluated on the validation set to determine the error estimate of the fold. This timeout was set to 150 minutes for classification and 15 minutes for feature search and evaluation in our experiments.\footnote{In preliminary experiments, only few models exceeded this timeout for the datasets studied here. \cite{snoek2011opportunity} presents a promising approach for using runtime predictions in the expected improvement calculation to automatically drive the search away from excessively expensive models. We plan to incorporate this approach into future versions of Auto-WEKA.}
{For each dataset, we ran Auto-WEKA with each hyperparameter optimization algorithm with a total time budget of 30 hours. For each method, we performed 25 runs of this process with different random seeds and then -- in order to simulate parallelization on a typical workstation -- used bootstrap sampling to repeatedly select 4 random runs and report the performance of the one with best cross-validation performance.}

{In early experiments, we observed a few cases in which Auto-WEKA's SMBO method picked hyperparameters that had excellent training performance, but turned out to generalize poorly. To enable Auto-WEKA to detect cases of overfitting, we partitioned its training set into two subsets: 70\% for use inside the SMBO method, and 30\% of validation data that is only looked at after the SMBO method has finished.}

\subsection{Algorithm Selection and CASH: Baseline Methods}\label{sec:baselines}

Auto-WEKA aims to aid non-expert users of machine learning techniques.
The simplest approach for selecting a classifier that is widely adopted amongst non-experts is to use a classifier merely based on its popularity or intuitive appeal, without any empirical consideration of alternatives. 
To quantify the difference such a choice can make, we consider the 39 WEKA classifiers (each with default hyperparameter settings) for each dataset, train each on the training set, and measure its accuracy on the test set. 
For each dataset, the second and third columns in Table~\ref{table:performances} present the best
and worst ``oracle performance'' of these classifiers on the test set.
We observe that the gap between the best and worst classifier was huge, \eg{} misclassification rates of 4.93\% \vs{} 99.24\% on the Dorothea dataset. 
Even when the set of classifiers was restricted to a few popular ones (we considered neural networks, random forests, SVMs, AdaBoost, C4.5 decision trees, logistic regression, and KNN), this gap still exceeded 20\% on 14 out of the 21 datasets. Furthermore, there was no single method that achieved good performance across all datasets: every method was at least 22\% worse than the best for at least one data set.
We conclude that some form of algorithm selection is essential for achieving good performance.

\begin{table*}[t]
\vskip -0.05in
\caption{Performance on both 10-fold cross-validation and test data. Random Grid Search was run once for an average of 400 hours per dataset; for SMAC and TPE, we performed 25 runs of 30 hours each. We report results as median percent error rate across 100\,000 bootstrap samples simulating 4 parallel runs. Ex-Def is deterministic. Test error rates are determined by training the selected model/hyperparameters on the entire 70\% training data and computing the accuracy on the previously unused 30\% test data. Boldface indicates the lowest error within a block of comparable methods. SC denotes correlation coefficients (see Section \ref{sec:exp_result_test}).}
  \label{table:performances}
  \setlength{\tabcolsep}{1.8pt}
    {\scriptsize
  \vspace*{-0.1in} 
\centering
  \begin{center}
\begin{sc}
\begin{tabular}{l@{\hskip 1.6em}cccccccccccccccccc}\toprule
\multirow{3}{*}{\textbf{Dataset}} & \multicolumn{5}{c}{\textbf{Oracle Perf. (\%)}} & $\quad$  &\multicolumn{4}{c}{\textbf{10-Fold C.V.\ Performance (\%)}} & $\quad$ &\multicolumn{4}{c}{\textbf{Test Performance (\%)}} &  &\multicolumn{2}{c}{\textbf{SC}} \\\cmidrule{2-6}\cmidrule{8-11}\cmidrule{13-16}\cmidrule{18-19}
&\multicolumn{2}{c}{Ex-Def} & & \multicolumn{2}{c}{Rand. Grid} && \multirow{2}{*}{Ex-Def} & \multirow{2}{*}{\parbox{0.6cm}{Rand. Grid}} & \multicolumn{2}{c}{Auto-WEKA} && \multirow{2}{*}{Ex-Def} & \multirow{2}{*}{\parbox{0.6cm}{Rand. Grid}} & \multicolumn{2}{c}{Auto-WEKA} & & \multirow{2}{*}{TPE}  & \multirow{2}{*}{SMAC} \\
\cmidrule{2-3}\cmidrule{5-6}\cmidrule{10-11}\cmidrule{15-16}%
&Best&Worst&&Best&Worst&& & & TPE & SMAC &&&  & TPE & SMAC & &  & 
\\

\midrule
Dexter & 7.78 & 52.78 &  & \textbf{5.00} & 58.33 &  & 10.20 & 7.48 & 9.90 & \textbf{5.48} &  & 8.89 & \textbf{5.00} & 9.44 & 7.22 &  & 0.82 & 0.25 \\
GermanCredit & 26.00 & 38.00 &  & \textbf{25.00} & 63.67 &  & 22.45 & 22.45 & 21.43 & \textbf{19.59} &  & \textbf{27.33} & \textbf{27.33} & 27.67 & 28.33 &  & 0.31 & 0.20 \\
Dorothea & \textbf{4.93} & 99.24 &  & \textbf{4.93} & 99.24 &  & 6.03 & 6.03 & 6.93 & \textbf{5.52} &  & 6.96 & 6.96 & 6.96 & \textbf{6.38} &  & 0.95 & 0.40 \\
Yeast & 40.00 & 68.99 &  & \textbf{37.53} & 68.99 &  & 39.43 & 38.87 & \textbf{35.03} & 36.27 &  & \textbf{40.45} & 40.90 & 41.12 & \textbf{40.45} &  & 0.36 & 0.49 \\
Amazon & \textbf{28.44} & 99.33 &  & \textbf{28.44} & 99.33 &  & \textbf{43.94} & \textbf{43.94} & 48.43 & 48.30 &  & \textbf{28.44} & \textbf{28.44} & 37.56 & 37.56 &  & 0.92 & 0.97 \\
Secom & 7.87 & 14.26 &  & \textbf{7.66} & 40.64 &  & 6.25 & 6.12 & 6.25 & \textbf{5.34} &  & 8.09 & 8.30 & \textbf{7.87} & \textbf{7.87} &  & -0.10 & -0.56 \\
Semeion & 8.18 & 92.45 &  & \textbf{6.08} & 92.45 &  & 6.52 & 6.52 & 6.91 & \textbf{4.86} &  & 8.18 & 8.18 & 8.18 & \textbf{5.03} &  & 0.84 & 0.73 \\
Car & 0.77 & 29.15 &  & \textbf{0.19} & 31.66 &  & 2.71 & 1.54 & 0.94 & \textbf{0.71} &  & 0.77 & 0.19 & \textbf{0.00} & 0.58 &  & 0.12 & 0.75 \\
Madelon & \textbf{17.05} & 50.26 &  & \textbf{17.05} & 51.03 &  & 25.98 & 24.26 & 24.26 & \textbf{20.87} &  & 21.38 & \textbf{20.77} & \textbf{20.77} & 21.15 &  & 0.44 & 0.43 \\
KR-vs-KP & 0.31 & 48.96 &  & \textbf{0.21} & 51.04 &  & 0.89 & 0.70 & 0.45 & \textbf{0.32} &  & \textbf{0.31} & 0.52 & 0.52 & \textbf{0.31} &  & 0.22 & 0.32 \\
Abalone & 73.18 & 84.04 &  & \textbf{72.55} & 89.23 &  & 73.33 & 72.45 & 72.20 & \textbf{71.76} &  & 73.18 & 72.79 & \textbf{72.71} & 73.02 &  & 0.15 & 0.10 \\
Wine Quality & 36.35 & 60.99 &  & \textbf{36.08} & 81.62 &  & 38.94 & 37.28 & 35.94 & \textbf{34.74} &  & 37.51 & 36.08 & \textbf{33.56} & 33.70 &  & 0.73 & 0.85 \\
Waveform & 14.27 & 68.80 &  & \textbf{14.20} & 68.80 &  & 12.73 & 12.73 & 12.57 & \textbf{11.71} &  & 14.40 & 14.40 & \textbf{14.20} & 14.40 &  & 0.36 & 0.26 \\
Gisette & 2.52 & 50.91 &  & \textbf{2.38} & 50.91 &  & 3.62 & 3.27 & 3.70 & \textbf{2.42} &  & 2.81 & 2.38 & 2.57 & \textbf{2.24} &  & 0.69 & 0.79 \\
Convex & \textbf{25.96} & 50.00 &  & \textbf{25.96} & 50.57 &  & 28.68 & 28.50 & 29.04 & \textbf{24.70} &  & 25.96 & 26.76 & 25.45 & \textbf{22.05} &  & 0.98 & 0.84 \\
\addlinespace[\interrowspace]
CIFAR-10-Small & 65.91 & 90.00 &  & \textbf{64.54} & 90.00 &  & 66.59 & 65.11 & 57.97 & \textbf{57.76} &  & 65.91 & 64.54 & 56.65 & \textbf{55.93} &  & 0.93 & 0.80 \\
MNIST Basic & 5.19 & 88.75 &  & \textbf{3.79} & 88.75 &  & 5.12 & 4.00 & 13.64 & \textbf{3.64} &  & 5.19 & 3.79 & 18.03 & \textbf{3.56} &  & 1.00 & 0.87 \\
Rot. MNIST + BI & 63.14 & 88.88 &  & \textbf{57.28} & 90.96 &  & 66.15 & 59.75 & 73.04 & \textbf{59.61} &  & 63.14 & 58.16 & 69.86 & \textbf{55.84} &  & 0.50 & 0.95 \\
Shuttle & 0.0138 & 20.8414 &  & \textbf{0.0069} & 20.8414 &  & 0.0328 & 0.0263 & \textbf{0.0230} & \textbf{0.0230} &  & 0.0138 & 0.0276 & \textbf{0.0069} & \textbf{0.0069} &  & 0.60 & 0.73 \\
KDD09-Appentency & 1.74 & 6.97 &  & \textbf{1.64} & 54.08 &  & 1.88 & 1.88 & 1.88 & \textbf{1.75} &  & 1.75 & 1.77 & \textbf{1.74} & \textbf{1.74} &  & 0.89 & 1.00 \\
CIFAR-10 & \textbf{64.27} & 90.00 &  & \textbf{64.27} & 90.00 &  & 65.54 & 65.54 & 66.68 & \textbf{63.21} &  & 64.27 & 64.27 & 64.80 & \textbf{62.39} &  & 0.33 & 0.69 \\

      \bottomrule
    \end{tabular}
    \end{sc}
  \end{center}
  }
  \vspace*{-0.2in} 
\end{table*}

A straight-forward algorithm selection method is to 
perform exhaustive 10-fold cross-validation on the training set and to return the classifier with the smallest average misclassification error across folds. We will refer to this method applied to the set of 39 WEKA classifiers as \emph{Ex-Def}; it is the best choice that can be made among the 39 WEKA classifiers (with their default hyperparameters) based on an exhaustive cross-validation and will serve as a baseline to compare an optimal solution to the algorithm selection problem in WEKA to our solution for the CASH problem in WEKA.

More experienced users of machine learning algorithms would not only select between a fixed set of default algorithms, but would also consider different hyperparameter settings --- for example by performing a grid search over the hyperparameter space of a single classifier (as, e.g., implemented in WEKA\footnote{This is implemented in WEKA's CVParameterSelection class, see \url{weka.wikispaces.com/Optimizing+parameters}}). 
Since different learning algorithms perform well for different problems, more experienced users optimally would also want to consider different hyperparameter settings for more than one learning algorithm. 
Therefore, a stronger baseline we will use is an approach that --- in addition to the 39 WEKA default classifiers --- considers various hyperparameter settings for all of WEKA's 27 base classifiers. More precisely, this baseline considers a grid of hyperparameter settings for each of these 27 base classifier, and performs a random search~\cite{bergstra2012random} in the union of these grids (plus the 39 WEKA default classifiers). We refer to this baseline as \emph{Random Grid} and note that --- as an optimization approach in the joint space of algorithms and hyperparameter settings --- it is a simple CASH algorithm. 
We executed this Random Grid search for all our datasets in parallel, using 400 CPU hours on average per dataset (at least 120 hours for each).
Table~\ref{table:performances} (columns 4 and 5) shows the best and worst ``oracle performance'' on the test set across these Random Grid classifiers. Comparing these performances to the default performance, we note that in most cases even WEKA's best default algorithm could be improved by selecting better hyperparameter settings, sometimes rather substantially so: \eg{}, in the rotated MNIST with background images task, random grid search offered a 6\% improvement over Ex-Def. 
We conclude that choosing hyperparameter settings appropriately can lead to substantial differences in performance and that
because of this fact even a relatively simple (albeit computationally expensive) approach for CASH can outperform algorithm selection by itself.

\subsection{Results for Cross-Validation Performance}\label{sec:exp_result_cv}

With 786 hierarchical hyperparameters, Auto-WEKA's combined algorithm / hyperparameter space is very complex.
We now study how effectively SMAC and TPE could search this space to optimize 10-fold cross-validation performance, and compare their performance to that of the Ex-Def and Random Grid methods defined in the previous section.
The middle portion of Table~\ref{table:performances} reports the results.
{First, we note that random grid search over the hyperparameters of all base-classifiers yielded better results than Ex-Def in 14/21 cases (and tied in the remaining seven), which underlines the importance of not only choosing the right algorithm but of also setting its hyperparameters well. 
However, we note that this performance of random grid search is based on a very large time budget of an average of 400 CPU hours per dataset (650 CPU hours on average for each of the large datasets), making it a somewhat unrealistic alternative in practice. 
In contrast, Auto-WEKA was only run for $4\times 30$ CPU hours per dataset, but still yielded substantially better performance than 
random grid search, outperforming it in 20/21 cases (and performing worse in one\footnote{For the Amazon data set, WEKA's default implementation of support vector machines yielded a very strong error rate of 44\%, which was below that of SMAC's median performance. One of SMAC's 25 runs actually reached an error rate of 36\%, indicating that Auto-WEKA would be competitive given more time.}). 
Comparing the two Auto-WEKA variants, SMAC outperformed TPE in 19/21 cases, with one tie.
We note that sometimes Auto-WEKA's performance improvements over the other methods were substantial, with relative reductions of the cross-validation error rate exceeding 15\% in 12/21 cases.
}

We conclude that by searching Auto-WEKA's combined algorithm/hyperparameter space, we can effectively find models with much better cross-validation performance than by grid search over WEKA's base classifiers.

\subsection{Results for Test Performance}\label{sec:exp_result_test}

{The results just shown demonstrate that Auto-WEKA is effective at optimizing its given objective function}; however, this is not sufficient to allow us to conclude that it fits models that generalize well.  
As the hypothesis space of a machine learning algorithm grows, so does its potential for overfitting.
The use of cross-validation 
substantially increases Auto-WEKA's robustness
against overfitting, but since its hypothesis space is much larger than that of standard classification
algorithms, it is important to carefully study whether (and to what extent) overfitting poses a problem. 

{To evaluate generalization, we determined a combination of algorithm and hyperparameter settings %
$A_{\lambda}$ by running Auto-WEKA as before (cross-validating on the training set), trained $A_{\lambda}$ on the entire training set, and then evaluated the resulting model on the test set.
The right portion of Table~\ref{table:performances} reports the test performance obtained with all methods. 
Broadly speaking, similar trends held as for cross-validation performance: random grid search performed better than Ex-Def, and Auto-WEKA in turn outperformed random grid search. However, the performance differences were less pronounced: random grid search only yielded better results than Ex-Def in 9/21 cases, with 6/21 ties and 6/21 cases in which Ex-Def performed better. 
Auto-WEKA continued to outperform random grid search and Ex-Def in 15/21 cases, with 3 ties and 3 losses.} Notably, Auto-WEKA performed best on all of the 11 largest datasets; we attribute this to the fact that the risk of overfitting decreases with dataset size.
Sometimes, Auto-WEKA's performance improvements over the other methods were substantial, with relative reductions of the test error rate exceeding 15\% in 5/21 cases. 
{Comparing the different Auto-WEKA variants, SMAC outperformed TPE on 12 datasets and tied on 3, with TPE performing better on 6.}
{Even when compared to the unrealistic ``oracle best'' Random Grid classifier (which has access to the test set!), Auto-WEKA found algorithms/hyperparameters with a smaller error on 9/21 datasets, tied on 2, and was outperformed on the remaining 10. For the 10 largest datasets, it performed better in 8 cases, tied in 1, and only lost in 1.}

{As mentioned in our experimental setup, Auto-WEKA only used 70\% of its training set during the optimization of cross-validation performance, reserving the remaining 30\% for assessing the risk of overfitting. At any point in time Auto-WEKA's SMBO method keeps track of its \emph{incumbent} (the hyperparameter configuration with the lowest cross-validation error rate seen so far). After its SMBO method has finished, Auto-WEKA extracts a trajectory of these incumbents from it and computes their generalization performance on the withheld 30\% validation data. It then computes the Spearman rank coefficient between the sequence of training performances (evaluated by the SMBO method through cross-validation) and this generalization performance. 
The rightmost columns in Table~\ref{table:performances} (labelled SC) show the average correlation coefficient for each run of Auto-WEKA.
We note a general trend: as the absolute gap between cross-validation and test performance grows, this correlation coefficient decreases. The GermanCredit dataset is a good example where Auto-WEKA can signal that it only has low confidence in how well its chosen hyperparameters will generalize. We do note, however, that this weak signal has to be used with caution: there is no guarantee that large correlation coefficients yield a small gap and vice versa.}

\subsection{{Classifiers Selected by Auto-WEKA}}\label{sec:class_selected}

\begin{figure}[t]
  \begin{center}
  {\scriptsize Selected Classifiers
  }
      \includegraphics[width=0.99\linewidth]{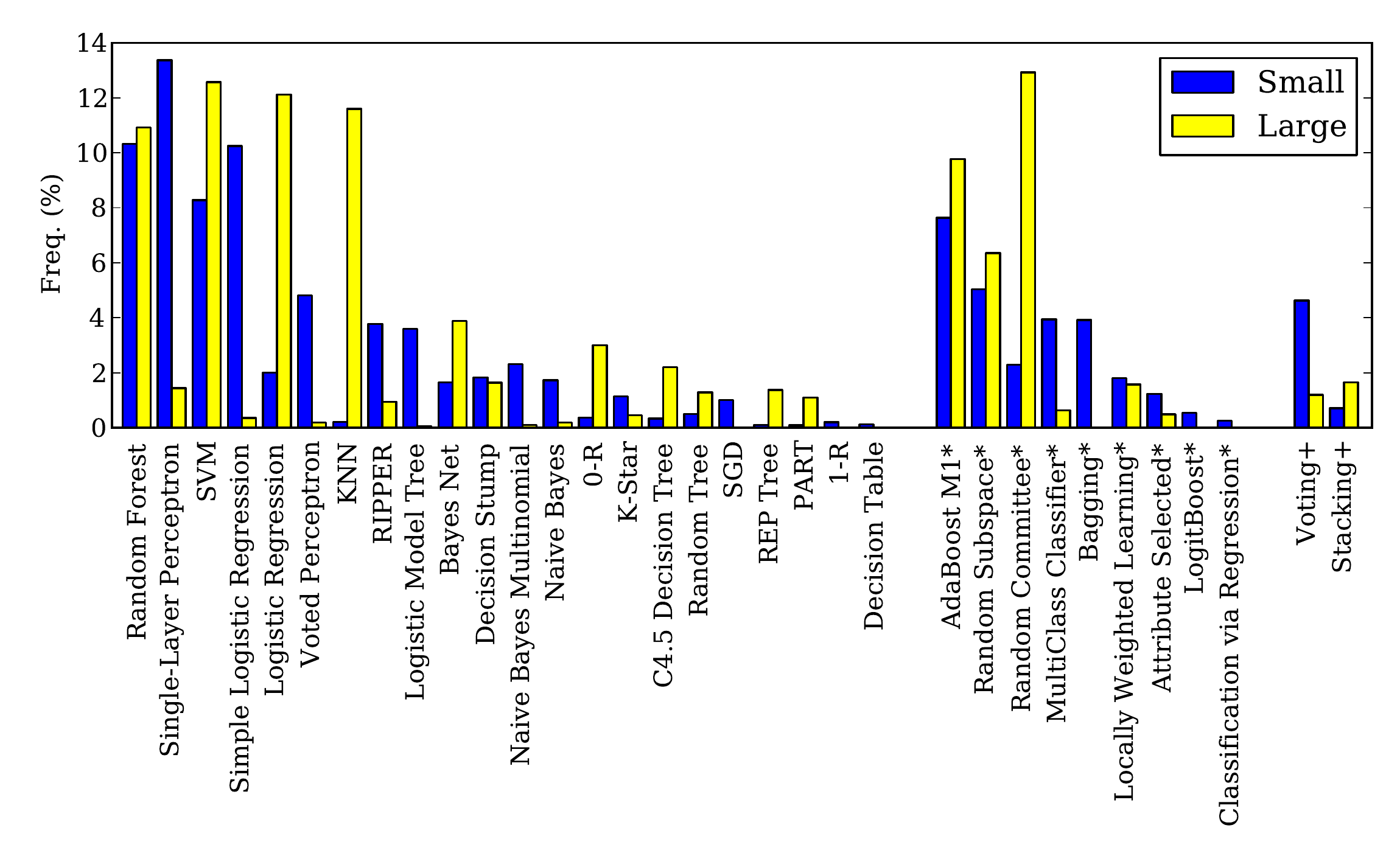}
  \end{center}
   \vspace*{-0.7cm}

  \caption{Distribution of chosen classifiers across the small and large datasets, aggregated across TPE, and SMAC, ranked on their frequency of being selected. Meta methods are marked by a $^*$ suffix, ensemble methods by a $^+$ suffix.}
  \label{fig:chosen}
  \vspace*{-0.1in} 
\end{figure}

\begin{figure}[t]
  \begin{center}
  \begin{tabular}{cc}%
    \begin{scriptsize}Selected Base Classifiers\end{scriptsize}
    &
    \begin{scriptsize}Feat. Search/Eval\end{scriptsize}
    \\
    
    \includegraphics[scale=0.295]{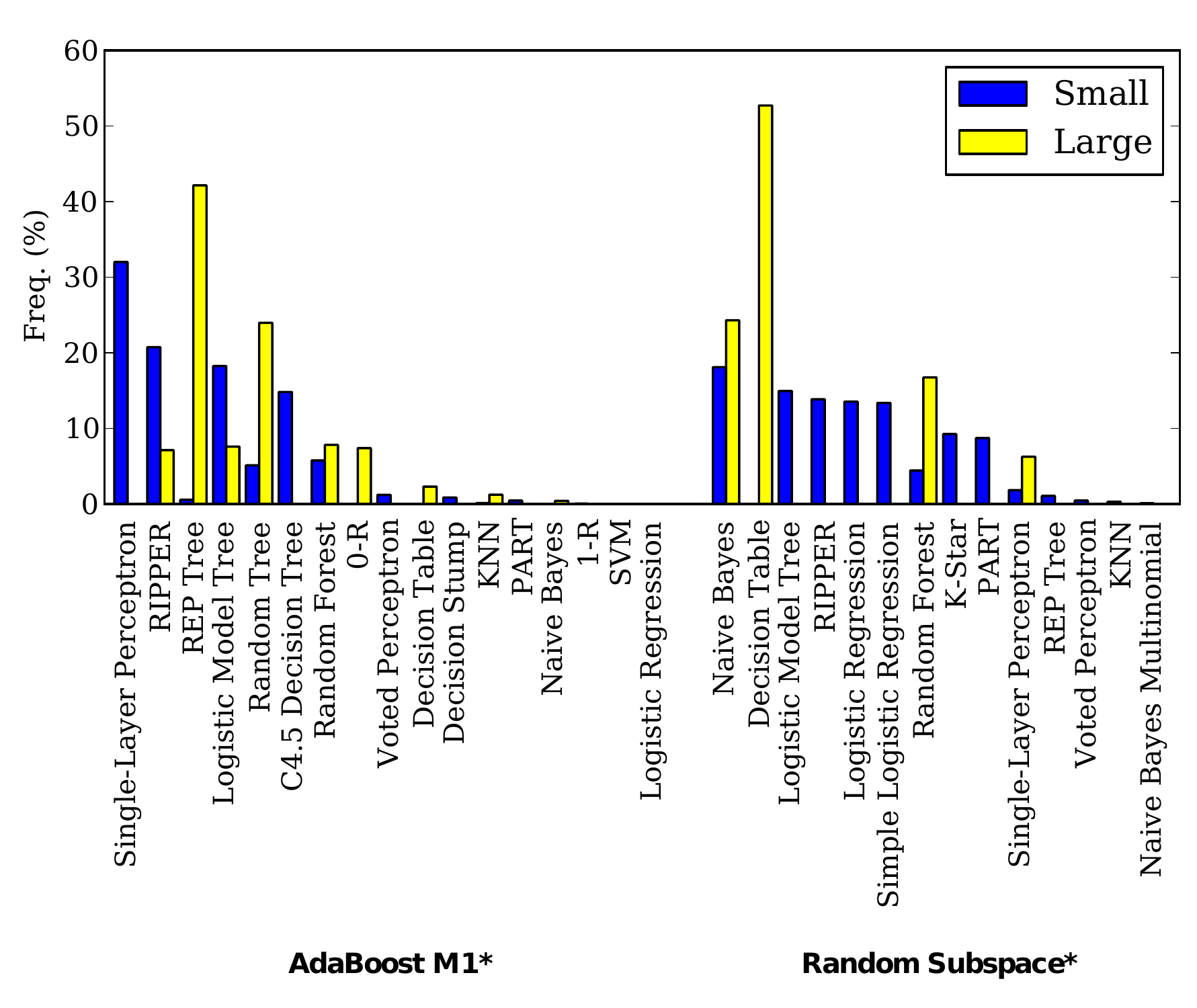}
    &
    \includegraphics[scale=0.295]{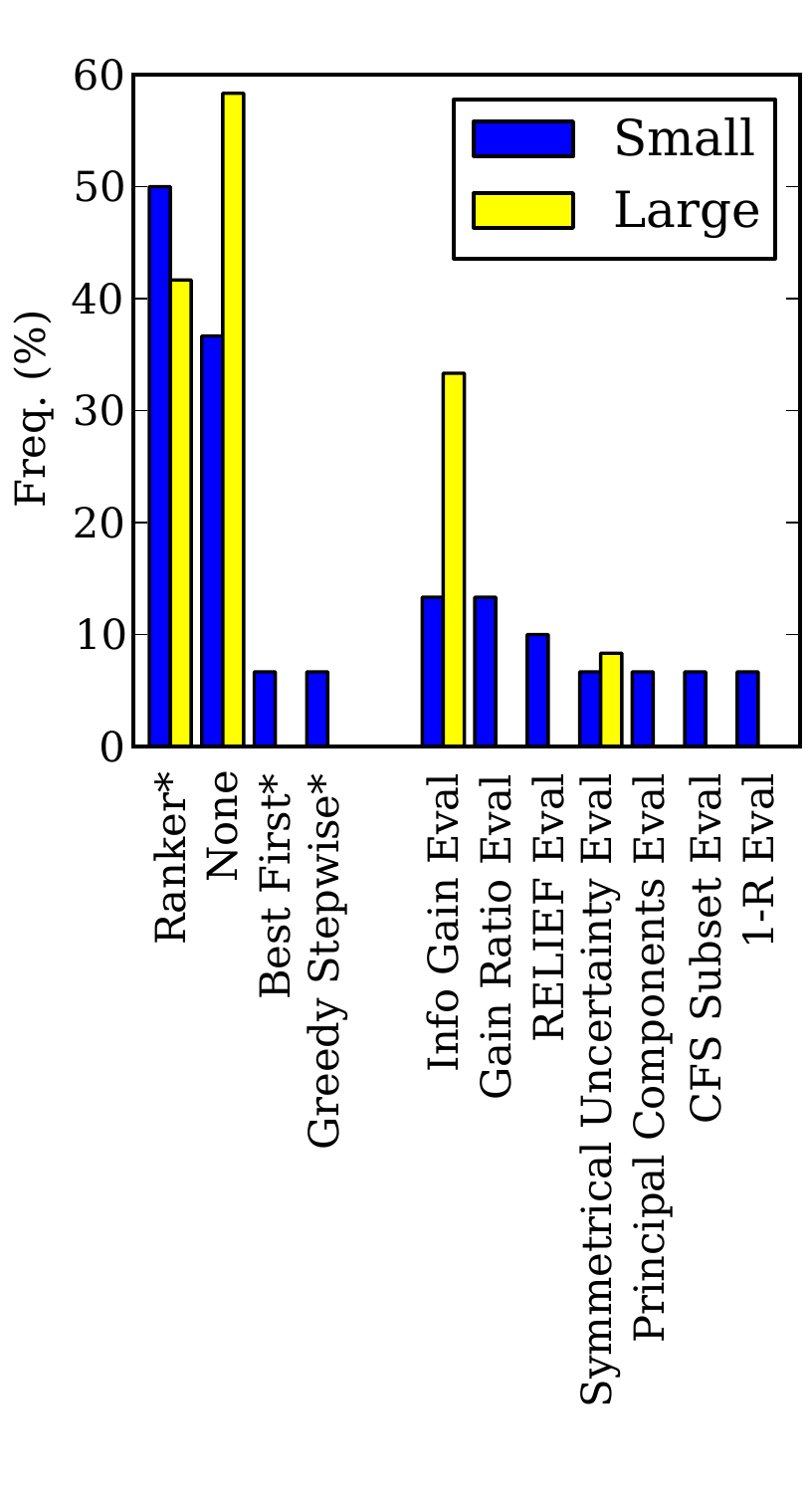}
   \end{tabular}
      
  \end{center}
  \vspace*{-0.6cm}
  \caption{Left: distribution of chosen base classifiers for the two most frequently selected meta methods: AdaBoostM1 and random subspace. Right: distribution of chosen feature search and evaluator methods. Both plots are aggregated across TPE and SMAC, ranked on their frequency of being selected; {\em None} indicates that no feature selection was performed.}
  \label{fig:chosen_meta}
  \vspace*{-0.1in} 
\end{figure}

Figure~\ref{fig:chosen} shows the distribution of classifiers chosen by our two Auto-WEKA variants (aggregated across runs and datasets - both TPE and SMAC produced similar results when considered individually). 
We note that no single classifier clearly dominated the others: the most frequently used classifiers (random forests, the single layer perceptron, and SVMs) were only selected in roughly 12\% of all cases each, and most classifiers were selected in at least a few percent of the cases. 
Furthermore, the selected methods differed considerably between the large and small datasets, demonstrating the need for dataset-specific methods; for example, the large datasets benefitted more from meta methods than the small ones.
A more detailed investigation of the top two meta-methods in Figure~\ref{fig:chosen_meta} (left) shows which base methods were chosen. Note that AdaBoostM1 frequently used the single layer perceptron on the small datasets, but never for the large ones, while the REP tree was highly popular for the large datasets. In the random subspace, the two most prominent methods were naive Bayes and the decision table.
It is interesting to note that these two methods, as well as the REP tree frequently selected by AdaBoost, were not often selected as a base classifier on their own.
This underlines the importance of searching Auto-WEKA's entire parameter space instead of, \eg{}, restricting one's attention to a small number of favourite base classifiers.

Figure~\ref{fig:chosen_meta} (right) provides a breakdown of the feature search and evaluation methods Auto-WEKA selected. 
Overall, it used these feature selection methods more often on the smaller datasets than on the larger ones, and if it 
did use a feature selection method it clearly favored the ranker method.
All feature evaluators were used with roughly the same frequency for small datasets; in contrast, if Auto-WEKA performed feature selection for a large dataset it clearly favored the information gain evaluator.
We note that Auto-WEKA's data-dependent choices (based on its internal cross-validation evaluation) allow it to use feature selection as a regularization method for small data sets, while at the same time using all features to construct more complex hypotheses for large datasets.

\section{Conclusion and future work}\label{sec:conclusion}

In this work, we have shown that the daunting problem of combined algorithm selection and hyperparameter optimization (short: CASH) can be solved by a practical, fully automated tool.
This is made possible by the use of recent
Bayesian optimization techniques that iteratively build models of the algorithm/hyperparameter landscape and leverage these models to identify new points in the space that deserve investigation.

We built a tool, Auto-WEKA, that utilizes the full range of classification algorithms in WEKA and makes it easy for non-experts to build high-quality classifiers for given application scenarios.
An extensive empirical comparison on 21 prominent datasets showed that Auto-WEKA often outperformed 
standard algorithm selection/hyperparameter optimization methods, especially on large datasets. 
We empirically compared two different optimizers for searching Auto-WEKA's 786-dimensional parameter
space and in the end recommend an Auto-WEKA variant based on the Bayesian optimization method SMAC~\cite{hutter2011smac}. We have written a freely-downloadable software package to make Auto-WEKA easy for end-users to access; it is available at \url{www.cs.ubc.ca/labs/beta/Projects/autoweka/}.

We see several promising avenues for future work.
First, Auto-WEKA still shows larger improvements in cross-validation performance than on test data, suggesting the investigation of more sophisticated methods for detecting and avoiding overfitting than our simple correlation-based approach.  
Second, we see potential value in extending our current approach to allow parameter sharing between classifiers used within ensemble methods, likely increasing their chance of being selected by Auto-WEKA.
Finally, we could use our approach as an inner loop for training
ensembles of machine learning algorithms
by iteratively adding algorithms with maximal marginal contribution
(this idea is conceptually related to the Hydra approach for constructing
algorithm selectors~\cite{Hydra10}).

\vspace*{-0.2cm}
\renewcommand{\baselinestretch}{0.9}

\bibliography{short,refs}
\bibliographystyle{abbrv}

\end{document}